\documentclass[10pt,twocolumn,letterpaper]{article}

\usepackage{iccv}              % To produce the CAMERA-READY version
\usepackage{multirow}
\usepackage{amsfonts} 
\usepackage{float}
\usepackage{comment}
\usepackage{subcaption}
\usepackage{booktabs}
\usepackage{adjustbox}
\usepackage{wrapfig}
\usepackage[table]{xcolor}
\usepackage{bm} 
\usepackage[accsupp]{axessibility}  % Improves PDF readability for those with disabilities.

\allowdisplaybreaks[2]

\definecolor{iccvblue}{rgb}{0.21,0.49,0.74}
\usepackage[pagebackref,breaklinks,colorlinks,allcolors=iccvblue]{hyperref}

\def\paperID{6111} % *** Enter the Paper ID here
\def\confName{ICCV}
\def\confYear{2025}

\title{DAP-MAE: Domain-Adaptive Point Cloud Masked Autoencoder for Effective Cross-Domain Learning}

\author{
Ziqi Gao\textsuperscript{1,3,4,*},
Qiufu Li\textsuperscript{2,3,4,*},
Linlin Shen\textsuperscript{2,3,4,\dag} \\
\textsuperscript{1}School of Computer Science \& Software Engineering, Shenzhen University\\
\textsuperscript{2}School of Artificial Intelligence, Shenzhen University\\
\textsuperscript{3}National Engineering Laboratory for Big Data System Computing Technology, Shenzhen University\\
\textsuperscript{4}Guangdong Provincial Key Laboratory of Intelligent Information Processing, Shenzhen University\\
{\tt\small gaoziqi2023@email.szu.edu.cn, \{liqiufu, llshen\}@szu.edu.cn}
}
\begin{document}

\twocolumn[{%
\maketitle
\vspace{-15pt}
\begin{figure}[H]
\hsize=\textwidth
\centering
\includegraphics[width=17.5cm]{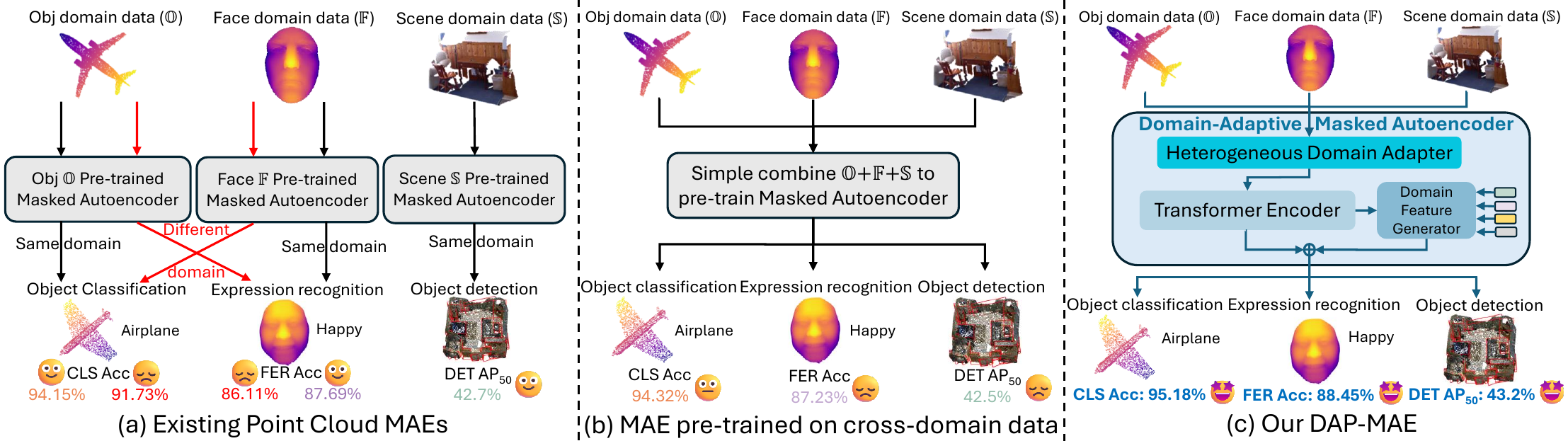}
\vspace{-15pt}
\caption{(a) \textbf{Existing point cloud MAEs} are typically pre-trained and fine-tuned within a single domain, leading to performance degradation when applied to other tasks in different domains. 
(b) Pre-training MAE directly on cross-domain data could also lead to performance decrease due to the misinterpretation of out-of-domain information.
%(b) Simply pre-training MAE on various datasets from different domains together does not obviously improve the performance of the downstream task and may even degrade it by misinterpreting out-of-domain information. 
(c) \textbf{Our DAP-MAE} 
collaboratively learns from point cloud data across various domains using the heterogeneous domain adapter and adopts the domain feature generator to extract diverse domain features, allowing the model to achieve high performance across multiple tasks with one pre-training in a single modality.}
%enables mixed-domain pre-training and simultaneous fine-tuning across multiple domains without performance degradation.}
\vspace{-5pt}
\label{fig:1}
\end{figure}
}]
\let\thefootnote\relax
\footnotetext{\textsuperscript{*}Equal contribution.}
\footnotetext{\textsuperscript{\dag}Corresponding author.}
%\vspace{-40pt}
\begin{abstract}
%\vspace{10pt}
Compared to 2D data, the scale of point cloud data in different domains available for training, is quite limited. Researchers have been trying to combine these data of different domains for masked autoencoder (MAE) pre-training to leverage such a data scarcity issue. However, the prior knowledge learned from mixed domains may not align well with the downstream 3D point cloud analysis tasks, leading to degraded performance. To address such an issue, we propose the \textbf{Domain-Adaptive Point Cloud Masked Autoencoder (DAP-MAE)}, an MAE pre-training method, to adaptively integrate the knowledge of cross-domain datasets for general point cloud analysis. In DAP-MAE, we design a heterogeneous domain adapter that utilizes an adaptation mode during pre-training, enabling the model to comprehensively learn information from point clouds across different domains, while employing a fusion mode in the fine-tuning to enhance point cloud features. Meanwhile, DAP-MAE incorporates a domain feature generator to guide the adaptation of point cloud features to various downstream tasks. With only one pre-training, DAP-MAE achieves excellent performance across four different point cloud analysis tasks, reaching 95.18\% in object classification on ScanObjectNN and 88.45\% in facial expression recognition on Bosphorus. The code will be released at \url{https://github.com/CVI-SZU/DAP-MAE}
\end{abstract}    
\vspace{-15pt}
\section{Introduction}
Point cloud analysis is gaining increasing attention in the fields of autonomous driving, robotics, and augmented/virtual reality, etc. It involves tasks such as point cloud-based object classification, part segmentation, facial expression recognition, and object detection, all of them require training feature extraction models that can effectively represent the geometric information of 3D point clouds. However, compared to 2D data, the collection and annotation of 3D point clouds still require substantial resources, resulting in small-scale labeled point cloud datasets, which limits the performance of supervised learning methods \cite{pointnet,pointnet++,pointmlp}. Self-supervised learning \cite{pointcontrast,maskpoint,point-mae} offers a promising alternative for the model training, utilizing unlabeled 3D point clouds and pre-training strategies to learn the features.

The self-supervised learning method based on masked autoencoders (MAE) can fully leverage the geometric information of 3D point clouds, enhancing the performance of downstream tasks, and various MAE-based methods for 3D point cloud analysis have been developed, such as Point-MAE \cite{point-mae}, PiMAE \cite{pimae}, and 3DFaceMAE \cite{3dfacemae}, etc. As \cref{fig:1} shows, when these methods are applied to downstream tasks including object classification, facial expression recognition, and indoor object detection, they require separate pre-training on the data from the same domain as the tasks, which leads to redundant consumption of pre-training and does not fully utilize the existing point cloud data from different domains. However, if the MAE is simply pre-trained using 3D point cloud data from multiple different domains, the downstream tasks may interpret the out-of-domain information learned by the feature model as interference noises, thereby reducing the task performance.
In this paper, to comprehensively utilize the 3D point cloud data from different domains, we designed a domain-adaptive point cloud masked autoencoder (DAP-MAE). DAP-MAE employs a heterogeneous domain adapter (HDA) and a domain feature generator (DFG) to collaboratively learn point cloud data from various domains, enabling the model to adaptively perform different downstream tasks with just one pre-training in a single modality, as \cref{fig:1}(c) shows.
Specifically, during the pre-training, the HDA employs an adaptation mode to process masked point clouds from different domains using three parallel MLPs, enabling the model to collaboratively learn cross-domain information. During the fine-tuning, the HDA utilizes a fusion mode to integrate the point cloud data output from the three MLPs and enhance the point cloud features extracted by the MAE. Meanwhile, the DFG is pre-trained to extract relevant domain features of the point cloud using a contrastive loss to guide the adaptation of the point cloud features to various downstream tasks in the fine-tuning.
Our main contributions are summarized as follows:
% Specifically, during the pre-training stage, a batch of point clouds from various domain datasets is processed by the dynamic domain adapter, which performs domain selection to route data into their corresponding domain-specific layers. This mechanism enables the model to learn distinct point cloud distributions across different domains. To further enhance the model's ability to distinguish point clouds from different domains, we concatenate a domain token to the point cloud features of each sample and input them into the Uni-Task Masked Autoencoder. 
% Simultaneously, a token generator is employed to produce domain tokens specific to each sample, and domain contrastive loss is applied across multiple domains within each batch to bring intra-domain samples closer together while pushing inter-domain samples further apart. During fine-tuning, the pre-trained dynamic domain adapter is frozen and transferred to downstream tasks. For each task, features from the corresponding domain-specific layer serve as main features, while those from other layers act as auxiliary features. A trainable MLP generates scaling factors from the original point cloud features to refine the main features. Our main contributions are summarized as follows:
\begin{itemize}
    \item For the first time, we propose a domain-adaptive point cloud masked autoencoder, DAP-MAE, that can collaboratively utilize point clouds from different domains for various downstream tasks with just one pre-training in a single modality.
    \item In DAP-MAE, we design a heterogeneous domain adapter to address collaborative learning on cross-domain data and a domain feature generator extracting various domain features to guide the downstream tasks.
    \item The experiments demonstrate that DAP-MAE achieves SOTA results in various point cloud analysis tasks including object classification, facial expression recognition, part segmentation, and object detection, surpassing or comparable to cross-modal methods.
\end{itemize}

\section{Related Work}
\subsection{3D point cloud analysis with deep learning}
Deep learning-based point cloud analysis has been applied to various downstream tasks. 
Early point cloud analysis methods focused on object analysis tasks like object classification and part segmentation \cite{pointnet,pointnet++,dgcnn,pointmlp,pointcnn,pointnext,pct}. PointNet \cite{pointnet} pioneered a direct approach to consuming unordered point sets using symmetric functions to capture global features, and its extension, PointNet++ \cite{pointnet++} introduced a hierarchical set abstraction that captures local information. 
Built on deeper research in object analysis tasks, some methods have shifted their focus to 3D face analysis tasks \cite{pointface,pointfaceformer,drfer,pointsurface}. Among these, PointFace \cite{pointface} introduced the first 3D face recognition method based on PointNet++ and also proposed a transformer-based variant called PointFaceFormer \cite{pointfaceformer}. Meanwhile, DrFER \cite{drfer} put forward the first 3D facial expression recognition method.
At the same time, point cloud analysis has also advanced toward 3D scene understanding \cite{votenet,3detr,pointrcnn,voxelnet}. 3DETR \cite{3detr} employs a fully transformer-based design to capture both local geometry and global context for enhancement of 3D object detection accuracy.
Despite their strong performance in individual tasks, these methods differ in structure, a unified model is thus required for all the downstream tasks. Meanwhile, by integrating the power of Transformers with self-supervised learning on unlabeled point clouds, Point-MAE \cite{point-mae}—the first MAE-based method under a standard transformer framework—has emerged as a promising way to address this limitation.
\begin{figure*}[t]
    \centering
    \includegraphics[width=1\linewidth]{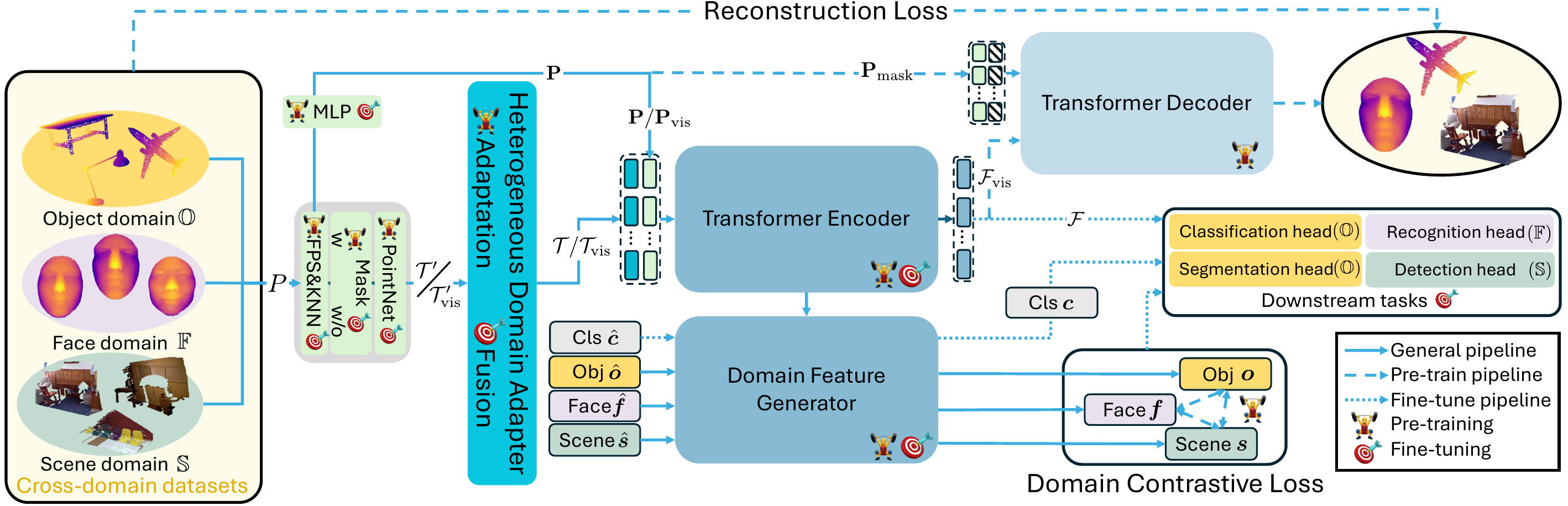}
    \vspace{-10pt}
    \caption{The pipeline of DAP-MAE. During the pre-training, the heterogeneous domain adapter (HDA) is set to adaptation mode to enhance the training of the Transformer-based MAE on cross-domain point cloud datasets, while the domain feature generator (DFG) is trained to extract domain features from the point clouds. In the fine-tuning phase, HDA is set to fusion mode, making comprehensive use of the feature extraction capabilities of the model trained on cross-domain data; meanwhile, the domain features extracted by DFG are used to guide the adaptation of point cloud features for downstream tasks.    }
    %A point cloud $P$ from cross-domain dataset is first divided into patches $\mathcal{P}$ using FPS and KNN, and its initial tokens $\mathcal T'$ and position embedding $\textbf{P}$ are respectively generated via PointNet and a MLP. 
    %During pre-training, after random masking, the visible initial tokens $\mathcal T'_{\text{vis}}$ are processed by our heterogeneous domain adapter (HDA) with adaptation mode to boost the collaboratively training of DAP-MAE on the cross-domain dataset, which is then input into Transformer encoder and output the feature $\mathcal F_{\text{vis}}$. The Transformer-based MAE is supervised by reconstruction loss and domain contrastive loss following the decoder and domain feature generator (DFG).     
    %The Transformer encoder extracts features used for two tasks: the decoder reconstructs masked patches via Chamfer distance loss, and the domain feature generator (DFG) extracts domain features. During fine-tuning, for a point cloud from a specialized domain, the HDA switches to fusion mode, and the transformer encoder extracts features that are decomposed into domain features and additional class features by DFG and then fed into downstream task heads.}
    \label{fig:3}
    \vspace{-10pt}
\end{figure*}
\subsection{MAE-based point cloud analysis}
With Point-MAE \cite{point-mae} pioneering the masked reconstruction paradigm in 3D vision, numerous generative approaches with cross-modal distillation have since been proposed. ACT \cite{act} transfers a Transformer pre-trained on other modalities to 3D point clouds and distills knowledges of other modalities into the point cloud Transformer. ReCon \cite{recon} unified generative modeling and contrastive learning through ensemble distillation guided by cross-modal teachers. Since cross-modal methods require pre-training on additional modalities, such a framework require high training costs, and highly rely on the performance of the teacher model. As a result, many approaches in 2024 have shifted back to improving single-modal strategies \cite{point-femae,patchmixing}. Also, some methods such as 3DFaceMAE \cite{3dfacemae} and Li \etal \cite{lihebeizi} have applied MAE-based methods to face analysis tasks. In the field of scene understanding, PiMAE \cite{pimae} and GDMAE \cite{gdmae} have adopted MAE-based methods for indoor and outdoor scene understanding tasks, respectively. However, existing methods have not thoroughly explored the relationships between different domains in current 3D point cloud datasets, resulting in many approaches performing well only on downstream tasks closely aligned with their pre-training datasets. %(\eg, ShapeNet for object point cloud tasks and FRGCv2 for face point cloud tasks). %Consequently, as Tab. \ref{tab:1} shows, their performance degrades significantly when handling cross-domain scenarios between pre-training and downstream datasets. This challenges models pre-trained on a single dataset to generalize effectively, so we propose DAUT that combines point cloud datasets from different domains for joint training to enhance performance across diverse downstream tasks.
% \begin{figure}[t]
%     \centering
%     \includegraphics[width=0.9\linewidth]{chart1.png}
%         \vspace{-5pt}
%     \caption{Performance of Point-MAE pre-trained on ShapeNet and FRGCv2, evaluated on object classification (ScanObjectNN OBJ\_BG split) and face expression recognition (Bosphorus), showing the impact of pre-training datasets on downstream tasks.}
%         \vspace{-5pt}
%     \label{fig:8}
% \end{figure}

\section{Preliminary}
The point cloud analysis addressed in this paper involves point clouds from three different data domains.
\begin{itemize}
  \item Object domain, $\mathbb O$. The tasks of 3D object classification and part segmentation utilize point cloud datasets from the object domain, such as ShapeNet \cite{shapenet}, ModelNet \cite{modelnet}, and ScanObjectNN \cite{scanojbectnn}.
  \item Face domain, $\mathbb F$. The tasks of 3D face recognition and facial expression recognition use point cloud datasets from the face domain, such as FRGCv2 \cite{frgc}, Bosphorus \cite{bosphorus}, and BU3DFE \cite{bu3dfe}.
  \item Scene domain, $\mathbb S$. The tasks of 3D object detection use point cloud datasets from the scene domain, such as ScanNet \cite{scannet} and S3DIS \cite{s3dis}.
\end{itemize}

The existing self-supervised point cloud analysis methods struggle to effectively utilize point cloud data from different domains. 
If the point clouds from different domains are used to pre-train feature extraction models, the performance of downstream tasks will significantly decline.
As \cref{fig:1}(a) shows, when we pre-train ReCon-SMC \cite{recon} on object data ($\mathbb O$) and then transfer it to object classification ($\mathbb O$), or pre-train it on face data ($\mathbb F$) and then transfer it to facial expression recognition ($\mathbb F$), their performance are 94.15\% and 87.69\%, respectively. However, when pre-training and the downstream task are conducted across different domains, such as pre-training on face data ($\mathbb F$) and transferring to object classification ($\mathbb O$), or pre-training on object data ($\mathbb O$) and transferring to facial expression recognition ($\mathbb F$), the performance drops to 91.73\% and 86.11\%, respectively.
Furthermore, directly combine point clouds from multiple domains together for the pre-training also struggles to improve the performance of downstream tasks.
As \cref{fig:1}(b) shows, when additional face data ($\mathbb F$) and scene data ($\mathbb S$) are sequentially added during the pre-training, the object classification ($\mathbb O$) performance only slightly improved by 0.17\%, resulting in an accuracy of 94.32\%. Moreover, the performance on facial expression recognition and object detection tasks even degrade to 87.23\% and 42.5\% respectively.

To fully exploit the information from data across different domains, this paper designs a domain-adaptive point cloud Masked Autoencoder, which requires only a single pre-training on point cloud data from multiple domains to achieve improved point cloud analysis accuracy on various downstream tasks.

\section{Method}
As \cref{fig:3} shows, in this section, we first introduce the overall pipeline of our DAP-MAE, followed by detail description of the heterogeneous domain adapter (HDA) and domain feature generator (DFG).

\subsection{The pipeline of DAP-MAE}
{\textbf{Datasets}.}\quad
To address the multiple point cloud analysis tasks including object classification, facial expression recognition, part segmentation, and object detection,
the proposed DAP-MAE is pre-trained on point clouds from three different domains, i.e., object $\mathbb O$, face $\mathbb F$, and scene $\mathbb S$,
while, in the fine-tuning phase, it is refined by the point clouds from the corresponding domain based on the specific downstream task.
\begin{figure}[t]
    \centering
    \includegraphics[width=1\linewidth]{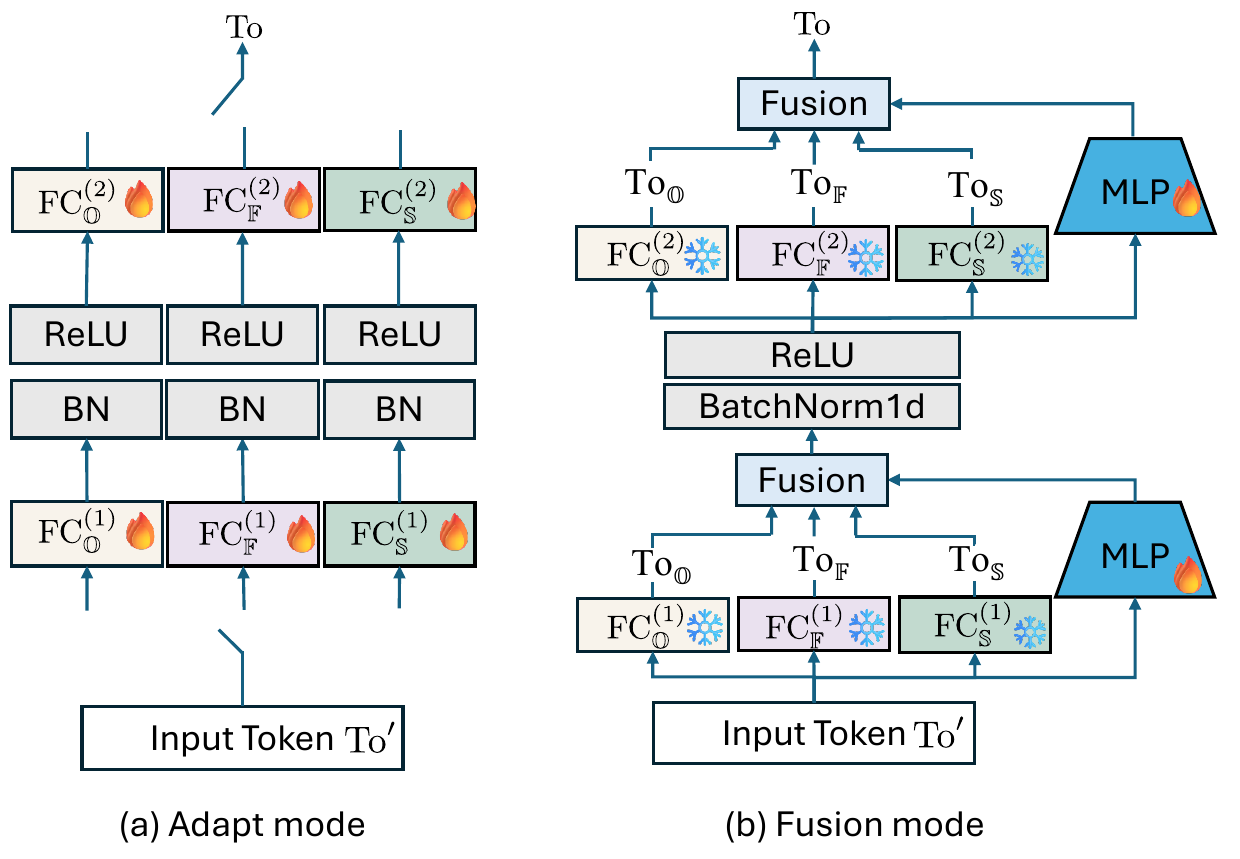}
    \vspace{-15pt}
    \caption{Illustration of the heterogeneous domain adapter (HDA), which consists of three parallel MLPs and two modes. (a) Adaptation mode. In the pre-training, HDA adaptively selects a MLP to process the input token based on its domain. (b) Fusion mode. In the fine-tuning, HDA freezes the parameters in the three MLPs and trains two additional MLPs to generate fusion coefficients, which are used to linearly fuse the output by the three MLPs.}
    \label{fig:4}
    \vspace{-15pt}
\end{figure}

\noindent{\textbf{Tokenization}.}\quad
For a point cloud $P$ from domain $\textbf{d}$, we first use farthest point sampling (FPS) and $k$-nearest neighbors (KNN) to partition it into a set of patches $\mathcal P$, where each patch in $\mathcal P$ contains a center point and $k$ neighboring points. In the pre-training, with a ratio of $m$, we randomly divide the patch set $\mathcal P$ into visible and masked parts,
\begin{align}
    \mathcal{P}_{\text{vis}}, \mathcal{P}_{\text{mask}}= \text{Mask} \circ \text{KNN} \circ \text{FPS}(P).
\end{align}
Before feeding into the Transformer-based MAE, we tokenize the patches using PointNet \cite{pointnet} and the proposed heterogeneous domain adapter (HDA),
\begin{align}
\mathcal T = [\mathcal T_{\text{vis}},\mathcal T_{\text{mask}}] = \text{HDA} \circ \text{PointNet} (\mathcal P_{\text{vis}},\mathcal P_{\text{mask}}),
\end{align}
and generate the position embedding using the center point of each patch by an MLP,
\begin{align}
\textbf{P} = [\textbf{P}_{\text{vis}}, \textbf{P}_{\text{mask}}] = \text{MLP}(\mathcal{P}_{\text{vis}}, \mathcal{P}_{\text{mask}}).
\end{align}

\noindent{\textbf{Reconstruction}.}\quad
In the pre-traing, we train the Transformer-based MAE on the visible data and reconstruct point clouds using a self-supervised manner.
The Transformer encoder extracts the feature of the point cloud from the visible $\mathcal{T}_{\text{vis}}$ and $\textbf{P}_{\text{vis}}$,
%which is added with $\textbf{P}_{\text{vis}}$ and fed into Transformer encoder in MAE, generating the feature of point cloud,
\begin{align}
\mathcal F_{\text{vis}} = \text{Encoder}(\mathcal T_{\text{vis}},\textbf{P}_{\text{vis}}),
\end{align}
and the decoder reconstructs the masked patches,
%In pre-training, the feature is used to predict the masked point patches by the Transformer decoder,
\begin{align}
    \mathcal{\hat{P}}_{\text{mask}}=\text{Decoder}(\mathcal{F}_{\text{vis}}, \textbf{P}_{\text{vis}}, \textbf{P}_{\text{mask}})).
\end{align}
The reconstruction in the MAE is supervised by a chamfer distance loss mutually comparing the patches in $\mathcal P_{\text{mask}}$ and $\hat{\mathcal P}_{\text{mask}}$,
\begin{align}
\nonumber
\mathcal L_\text{rec} &= \frac{1}{|\mathcal P_{\text{mask}}|}\sum_{\text{Pa}\in\mathcal P_{\text{mask}}}
                \min_{\text{Pa}'\in\hat{\mathcal P}_{\text{mask}}}
                \|\text{Pa}-\text{Pa}'\|_2^2\\
                &+\frac{1}{|\hat{\mathcal P}_{\text{mask}}|}\sum_{\text{Pa}\in\hat{\mathcal P}_{\text{mask}}}
                \min_{\text{Pa}'\in\mathcal P_{\text{mask}}}
                \|\text{Pa}-\text{Pa}'\|_2^2.
\end{align}

\noindent{\textbf{Domain contrasting}.}\quad
To extract domain features that guide the downstream tasks, we train the domain feature generator (DFG) in the pre-training to decompose the feature $\mathcal F_{\text{vis}}$ into the domain feature $\bm d$,
%and domain-invariant feature $\bm c$,
\begin{align}
\bm  d = \text{DFG}(\mathcal F_{\text{vis}}),
\end{align}
%where $\bm c$ is reserved for class feature in the fine-tuning.
In each iteration of the pre-training, DFG extracts the domain features $[\bm d_i]_{i=1}^B$ of a batch of cross-domain point clouds $[P_i]_{i=1}^B$ from domains $[\textbf d_i]_{i=1}^B$, $\textbf d_i \in \{\mathbb O, \mathbb F, \mathbb S\}$, and it is trained using a contrastive loss,
%DAPMAE receives a batch of cross-domain point clouds $[P_i]_{i=1}^B$ with domain labels $[\textbf d_i]_{i=1}^B$,
%and DFG takes a contrastive loss to constrain their generated domain features to train the parameters in DFG,
\begin{align}
\mathcal L_{\text{con}} = \sum_{i=1}^B\sum_{j=1,j\neq i}^B  l(\bm d_i, \bm d_j),
\end{align}
\begin{align}
l(\bm d_i, \bm d_j)=
\begin{cases}
1 - \text{cos}(\bm d_i, \bm d_j), & \textbf d_i = \textbf d_j \\
\max(0, \text{cos}(\bm d_i, \bm d_j) - a), & \textbf d_i \neq \textbf d_j, \end{cases}
\end{align}
and $a$ is a margin constant. 

\noindent{\textbf{Fine-tuning}.}\quad
During fine-tuning, for a point cloud $P$ from a specialized domain, the Transformer encoder extracts its features from the tokens $\mathcal T$, and the DFG utilizes these features to decompose domain features while additionally extracting class features $\bm c$ with class token,
\begin{align}
&\mathcal F = \text{Encoder}(\mathcal T, \textbf P),\\
&\bm c, \bm d = \text{DFG}(\mathcal F),
\end{align}
which are then fed into a downstream task head together, and the head is trained using its customized loss.

\subsection{Heterogeneous Domain Adapter}
The Heterogeneous Domain Adapter (HDA) consists of three parallel MLPs, which, in the pre-training and fine-tuning phases, process point cloud data using different modes. As \cref{fig:4} shows,
in the pre-training, HDA employs an adaptation mode, processing point cloud data from the three different domains using the different MLPs respectively, which synergistically leverages cross-domain data to enhance the feature extraction of encoder in MAE.
In the fine-tuning, HDA adopts a fusion mode, where the three MLPs simultaneously process the point cloud data and extract its features from their fused results.

\noindent\textbf{Adaptation mode.}\quad
Suppose a point could $P$ from domain $\textbf{d}, \textbf{d}\in\{\mathbb{O}, \mathbb F, \mathbb S\}$.
In the pre-training, for its visible patch set $\mathcal P_{\text{vis}}$, PointNet first produces its initial tokens,
\begin{align}
\mathcal T'_{\text{vis}} =\text{PointNet}(\mathcal P_{\text{vis}})= \{\text{PointNet}(\text{Pa}): \text{Pa}\in\mathcal P_{\text{vis}}\}.
\end{align}
Then, HDA generates its tokens using one of three parallel MLPs, $\{\text{MLP}_{\textbf d}: \textbf d \in \{\mathbb{O}, \mathbb F, \mathbb S\}\}$, according to the domain of $P\in\textbf d$,
\begin{align}
\mathcal T_{\text{vis}} &= \text{HDA}(\mathcal T'_{\text{vis}}) = \text{MLP}_{\textbf d}(\mathcal T'_{\text{vis}}) \nonumber\\
&= \{\text{To}=\text{MLP}_{\textbf d}(\text{To}'): \text{To}' \in \mathcal T'_{\text{vis}}\}.
\end{align}
In HDA, each of three MLPs contains two linear full connection (FC) layers, $\text{FC}_{\textbf d}^{(1)}$ and $\text{FC}_{\textbf d}^{(2)}$, with a BN and ReLU,
and for every initial token $\text{To}'$ in $\mathcal T'_{\text{vis}}$,
\begin{align}
\text{To} = \text{FC}_{\textbf d}^{(2)} \circ \text{ReLU} \circ \text{BN} \circ \text{FC}_{\textbf d}^{(1)}(\text{To}') \in \mathcal T_{\text{vis}}.
\end{align}

\noindent\textbf{Fusion mode.}\quad 
During the fine-tuning, for the different downstream task, we respectively fine-tune the feature extraction parameters of DAP-MAE and the downstream task heads on point cloud data from the same domain. For the point cloud $P$ from domain $\textbf d\in\{\mathbb O, \mathbb F, \mathbb S\}$, we generate all its initial tokens using FPS, KNN, and PointNet,
\begin{align}
\mathcal T' = \text{PointNet} \circ \text{KNN} \circ \text{FPS} (P).
\end{align}

{\color{black}In the fine-tuning, as \cref{fig:4} shows, HDA freezes its parameters and processes each initial token with the three parallel MLPs separately, and twice linearly fuses the MLP results.}
In the first fusion, for each initial token $\text{To}' \in \mathcal T'$ of point cloud $P\in \textbf d$, HDA processes it using the first three FCs in the three parallel MLPs,
\begin{align}
\text{To}_{\textbf d'}^{(1)} = \text{FC}^{(1)}_{\textbf d'}(\text{To}'), ~~\forall\textbf d'\in\{\mathbb O, \mathbb F, \mathbb S\}, \forall\text{To}' \in \mathcal T'.
\end{align}
In the fusion, the token output by the FC corresponding to the domain $\textbf{d}$ of downstream task is taken as the primary token, while the other tokens from FCs for other domains serve as auxiliary tokens,
\begin{align}
\text{To}^{(1)} = \text{To}_{\textbf d}^{(1)} + \sum_{\textbf d' \in \{\mathbb O, \mathbb F, \mathbb S\}\atop \textbf d' \neq \textbf d} \alpha_{\textbf d'} \text{To}_{\textbf d'}^{(1)},
\end{align}
where the two coefficients are calculated using one MLP on the initial token,
\begin{align}
\{\alpha_{\textbf d'}: \textbf d' \in\{\mathbb O, \mathbb F, \mathbb S\}, \textbf d' \neq \textbf d\} = \text{MLP}^{(1)}(\text{To}').
\end{align}
Then, in the second fusion, the token $\text{To}^{(1)}$ is first processed by ReLU and BN, and fused by the second three FCs of the three parallel MLPs with the similar way,
\begin{align}
\text{To}_{\textbf d'}^{(2)} &= \text{FC}^{(2)}_{\textbf d'}(\text{ReLU}(\text{BN}(\text{To}^{(1)}))), ~\textbf d'\in\{\mathbb O, \mathbb F, \mathbb S\},\\
\text{To} &= \text{To}_{\textbf d}^{(2)} + \sum_{\textbf d' \in \{\mathbb O, \mathbb F, \mathbb S\}\atop \textbf d' \neq \textbf d} \alpha_{\textbf d'} \text{To}_{\textbf d'}^{(2)},
\end{align}
where the two coefficients are calculated using another MLP on the token $\text{To}^{(1)}$,
\begin{align}
\{\alpha_{\textbf d'}: \textbf d' \in\{\mathbb O, \mathbb F, \mathbb S\}, \textbf d' \neq \textbf d\} = \text{MLP}^{(2)}(\text{To}^{(1)}).
\end{align}

\subsection{Domain Feature Generator}
The domain feature generator (DFG) extracts domain features from the point cloud features $\mathcal F_{\text{vis}}$ or $\mathcal F$ output by the Transformer encoder,
which are used to guide the training of downstream task heads during the fine-tuning phase.

In the DFG, a class token $\hat{\bm c}$ and three domain tokens $\{\hat{\bm o}, \hat{\bm f}, \hat{\bm s}\}$ are set.
During the pre-training, based on the domain $\textbf d$ of the original input point cloud $P$, one domain token $\hat{\bm d}\in \{\hat{\bm o}, \hat{\bm f}, \hat{\bm s}\}$ is selected and concatenated with the class token $\hat{\bm c}$.
Then, domain features are extracted from $\tilde{\mathcal F}$, $\tilde{\mathcal F} = \mathcal F_{\text{vis}}$ or $\mathcal F$ , using a cross-attention mechanism,
\begin{align}
&Q = \text{FC}_q([\hat{\bm c}, \hat{\bm d}]), K = \text{FC}_k(\tilde{\mathcal F}), V = \text{FC}_v(\tilde{\mathcal F}),\\
&[\bm c, \bm d] = \text{Attn}(Q,K,V) = \text{SoftMax}\bigg(\frac{QK^T}{\sqrt{d}}\bigg)V.
\end{align}
where $d$ is the dimension of the key vectors $K$ and $\bm c, \bm d$ are the class feature and domain feature extracted by DFG. To be clear, the class token $\hat{\bm c}$ are only supervised during the fine-tuning stage to extract class feature $\bm c$.

\begin{table*}[!t]
    \centering
    \footnotesize
    \setlength{\tabcolsep}{4pt}
        \caption{Object Classification results on the ScanObjectNN and Face Expression Recognition results on BU3DFE and Bosphorus. PM: pre-trained modality, \textit{PC}: point cloud, \textit{I}: image, \textit{T}: text.} 
        \vspace{-5pt}
    \begin{tabular}{c c c c c ||c c c}
        \hline
        \hline
        \multicolumn{5}{c||}{Object Classification}& \multicolumn{3}{c}{Facial Expression Recognition}\\
        \hline
        \multirow{2}{*}{Method} &\multirow{2}{*}{PM} &\multicolumn{3}{c||}{ScanObjectNN}&\multirow{2}{*}{Method}&\multirow{2}{*}{BU3DFE}&\multirow{2}{*}{BOS}\\
        \cline{3-5}
        && OBJ\_BG & OBJ\_ONLY &PB\_T50\_RS \\
        \hline
        \multicolumn{5}{c||}{\textit{Supervised Learning}}&\multicolumn{3}{c}{2D+3D}\\
        \cline{1-8}
        PointNet \cite{pointnet} $_{\text{CVPR'17}}$  &\textit{PC} &  73.3    & 79.2     & 68.0  &DA-CNN \cite{dacnn}&88.35 &-   \\
        PointNet++ \cite{pointnet++} $_{\text{NeruIPS'17}}$   & \textit{PC}   &  84.3    & 84.3     & 77.9&Jan \etal \cite{jan}& 88.54 &-   \\
        DGCNN \cite{dgcnn}  $_{\text{TOG'19}}$    &\textit{PC}&  82.6  & 86.2     & 78.1 &Zhu \etal \cite{intensity}&88.75 &-    \\
        PointCNN \cite{pointcnn} $_{\text{NeruIPS'18}}$&\textit{PC}  &   86.1    & 85.5     & 78.5 &CM-CNN \cite{CM-CNN}& 88.91 &85.16    \\
        SimpleView \cite{simpleview} $_{\text{ICML'21}}$ &\textit{PC}   &   -       & -        & 80.5±0.5 &FE3DNet\cite{fe3dnet}&89.05 &89.28   \\
        MVTN \cite{mvtn} $_{\text{ICCV'21}}$ & \textit{PC}& 92.6    & -        & 82.8 &FA-CNN \cite{FA-CNN}&89.11 &-\\
        PointMLP \cite{pointmlp} $_{\text{ICLR'22}}$ &\textit{PC}  &   -       & -        & 85.4±0.3 & Oyedotun \etal \cite{oyedotun}&89.31 &- \\
        PCT \cite{pct} $_{\text{CVM'21}}$& \textit{PC}&-       & -        & 83.4 &Jiao \etal \cite{Jiao}&89.72 &83.63  \\
        PointNeXt \cite{pointnext} $_{\text{NeruIPS'22}}$ &\textit{PC}    &    -       & -        & 87.7±0.4 &FFNet-M \cite{ffnet-m}&89.82 &87.65  \\
        P2P-HorNet \cite{p2p-hornet} $_{\text{NeruIPS'22}}$ &\textit{PC}    &   - & -      & 89.3 &AFNet-M \cite{afnet-m}&90.08& 88.31  \\
        SFR \cite{sfr} $_{\text{ICASSP'23}}$ &\textit{PC}  &   -& -& 87.8 & Cmanet \cite{cmanet} &90.24 &89.36 \\
        \cline{1-8}
        \multicolumn{5}{c||}{\textit{Self-Supervised Learning}}&\multicolumn{3}{c}{3D}\\
        \cline{1-8}
        Point-BERT \cite{point-bert} $_{\text{CVPR'22}}$&\textit{PC}& 87.43 & 88.11 & 83.07 &Jan \etal \cite{jan}& 81.83& - \\
        Point-MAE \cite{point-mae} $_{\text{ECCV'22}}$&\textit{PC}&90.02 & 88.29 & 85.18&CM-CNN \cite{CM-CNN}&80.11&77.82\\
        Point-M2AE \cite{point-m2ae} $_{\text{NeruIPS'22}}$&\textit{PC}&91.22& 88.81& 86.43&Zhen \etal \cite{zhen}& 84.50& -\\
        PointGPT-S \cite{pointgpt} $_{\text{NeruIPS'23}}$&\textit{PC}&91.60&90.00&86.90&FLM-CNN \cite{flm-cnn}&86.67& -\\
        PointDif \cite{pointdif} $_{\text{CVPR'24}}$&\textit{PC}&91.91 &93.29 &87.61&Zhu \etal \cite{intensity}&87.19& -\\
        Mamba3D+Point-MAE \cite{mamba3d} $_{\text{ACM MM'24}}$&\textit{PC}&93.12&92.08&88.20&FFNet-M \cite{ffnet-m}&87.28& 82.86\\
        Point-FEMAE \cite{point-femae} $_{\text{AAAI'24}}$&\textit{PC}&\textbf{95.18}& 93.29 &90.22&Oyedotun \etal \cite{oyedotun}& 84.72& -\\
        ReCon-SMC \cite{recon} $_{\text{ICML'23}}$ (baseline)&\textit{PC}&94.15& 93.12& 89.73&Yang \etal \cite{yang2015automatic}& 84.80& 77.50\\
        %\cellcolor{cyan!8} Point-MAE$^{\dagger}$ \cite{point-mae}&\cellcolor{cyan!8}\textit{PC}&\cellcolor{cyan!8}92.77 & \cellcolor{cyan!8}91.22  & \cellcolor{cyan!8}88.24&FE3DNet\cite{fe3dnet}&85.20 &83.55\\
        % \cellcolor{cyan!8}$\mathbb O+\mathbb F+\mathbb S$ Point-MAE+DAP-MAE&\cellcolor{cyan!8}\textit{PC}&\cellcolor{cyan!8}\color{blue}94.66  & \cellcolor{cyan!8} \color{blue}92.94 & \cellcolor{cyan!8}\color{blue}89.56&FE3DNet\cite{fe3dnet}&85.20 &83.55\\
        %\cellcolor{cyan!8} ReCon-SMC$^{\dagger}$ \cite{recon} (baseline)&\cellcolor{cyan!8}\textit{PC}&\cellcolor{cyan!8}94.32 & \cellcolor{cyan!8}93.12 & \cellcolor{cyan!8}89.90 &DA-CNN \cite{dacnn}&87.69 &-\\
         \textbf{DAP-MAE(Ours)}&\textit{PC}&\textbf{95.18} &\textbf{93.45} &\textbf{90.25} &FE3DNet\cite{fe3dnet}&85.20 &83.55\\
        \cline{1-5}
        ACT \cite{act} $_{\text{ICLR'23}}$&\textit{PC+I}&93.29&91.91&88.21&DA-CNN \cite{dacnn}&87.69 &-\\
        Joint-MAE \cite{joint-mae} $_{\text{IJCAI'23}}$&\textit{PC+I}&90.94&88.86&86.07&Cmanet \cite{cmanet}&84.03 &81.25\\
        I2P-MAE \cite{i2pmae} $_{\text{CVPR'23}}$&\textit{PC+I}&94.14&91.57&90.11&AFNet-M \cite{afnet-m}&86.97 &82.06\\
        TAP \cite{tap} $_{\text{ICCV'23}}$&\textit{PC+I}&90.36&89.50&85.67&DrFER \cite{drfer}&89.15&86.77\\
        ReCon-full \cite{recon} $_{\text{ICML'23}}$&\textit{PC+I+T}&95.18&93.63&90.63&\textbf{DAP-MAE(Ours)}&\textbf{89.83}&\textbf{88.45}\\
        \hline
        \hline
    \end{tabular}  
    \label{table1}

    \end{table*}

\begin{table*}[!t]
    \centering
    \footnotesize
    \caption{Results for object classification, facial expression recognition (FER), and object detection (DET). We compare Point-MAE and ReCon-SMC (both using same-domain pre-training and transfer) with their cross-domain version trained using the simple combination of different domain data, as well as our DAP-MAE. PM: pre-trained modality, \textit{PC}: point cloud. $\dagger$ denotes the reproduced results.}
    \vspace{-5pt}
    \begin{tabular}{ccccc||cc||cc}
    \hline
    \hline
    &&\multicolumn{3}{c||}{object classification}&\multicolumn{2}{c||}{FER}&\multicolumn{2}{c}{DET}\\
    \hline
    \multirow{2}{*}{Method}&\multirow{2}{*}{PM}&\multicolumn{3}{c||}{ScanObjectNN}&\multirow{2}{*}{BU3DFE}&\multirow{2}{*}{BOS}&\multicolumn{2}{c}{ScanNetV2}\\
    \cline{3-5}
    &&OBJ\_BG&OBJ\_ONLY&PB\_T50\_RS&&&AP$_{50}$&AP$_{25}$\\
    \hline
    \rowcolor{gray!10}Point-MAE \cite{point-mae} $_{\text{ECCV'22}}$&\textit{PC}&90.02 & 88.29 & 85.18&88.89&86.75&-&-\\
    ReCon-SMC \cite{recon} $_{\text{ICML'23}}$ (baseline)&\textit{PC}&94.15& 93.12& 89.73&89.13&87.69&42.7&63.8\\
    \hline
     \rowcolor{gray!10}$\mathbb O+\mathbb F+\mathbb S$ Point-MAE$^\dagger$ \cite{point-mae}&\textit{PC}&92.77&91.22&88.24&87.34&86.28&-&-  \\
     $\mathbb O+\mathbb F+\mathbb S$ ReCon-SMC $^\dagger$ \cite{recon} (baseline)&\textit{PC}&94.32&93.12&89.90&88.52&87.23&42.5&63.5\\
     \rowcolor{gray!30}\textbf{DAP-MAE (Ours)}&\textit{PC}&\textbf{95.18}&\textbf{93.45}&\textbf{90.25}&\textbf{89.83}&\textbf{88.45}&\textbf{43.2}&\textbf{64.0}    \\
    \hline
    \hline
    \end{tabular}
    \label{tab:my_label}
    \vspace{-10pt}
\end{table*}

\section{Experiments}
\subsection{Pre-training}
\noindent\textbf{Cross-domain dataset.}
The pre-training of DAP-MAE was conducted on a cross-domain point cloud dataset, which involve three datasets from different domains: ShapeNet ($\mathbb O$), FRGCv2 ($\mathbb F$), and S3DIS ($\mathbb S$). 
% ShapeNet \cite{shapenet} was captured from object ($\mathbb O$) domain, which contains 50,000 3D point clouds across 55 object categories. 
% The original FRGCv2 \cite{frgc} consists of 4,007 high-quality 3D face scans from 466 individuals with expression variations. In the pre-training, we utilized the enriched FRGCv2 \cite{gilani2018learning}, which contains 100K 3D face from 1K individuals. 
% S3DIS \cite{s3dis} consists of six large-scale indoor scenes from three different buildings, covering a total of 273 million points across 13 categories. Only the training split of S3DIS was used. 
For each point cloud in the cross-domain dataset, we uniformly sample its 4,096 points for the pre-training. More details about the cross-domain dataset will be provided in the supplementary material.

\subsection{Fine-tuning}
The pre-trained DAP-MAE was fine-tuned on five representative datasets covering diverse downstream tasks, including object classification ($\mathbb{O}$), few-shot learning ($\mathbb{O}$), part segmentation ($\mathbb{O}$), facial expression recognition ($\mathbb{F}$), and 3D object detection ($\mathbb{S}$). For each task, we selected widely used benchmark datasets. These include ScanObjectNN \cite{scanojbectnn} for object classification, ModelNet40 \cite{modelnet} for few-shot learning classification, ShapeNetPart \cite{shapenet} for part segmentation, BU-3DFE \cite{bu3dfe} and Bosphorus \cite{bosphorus} for facial expression recognition, and ScanNetV2 \cite{scannet} for indoor scene 3D object detection. More detailed descriptions and implementation settings can be found in the supplementary material.

\noindent\textbf{Object classification ($\mathbb O$).}
The left side of \cref{table1} shows the classification accuracy (\%) on ScanObjectNN. 
Our DAP-MAE outperforms the latest single-modal self-supervised learning approaches across all three protocols and exceeds the performance of most cross-modal methods.
Additionally, in \cref{tab:my_label}, we compare Point-MAE and ReCon-SMC, which are both pre-trained and transferred on the same domain with their counterparts pre-trained on the cross-domain dataset. The results show that simply adding cross-domain data only gains slight improvements in object classification for ReCon-SMC, whereas our improved DAP-MAE significantly surpasses the single-modal baseline ReCon-SMC by 1.03\% and its cross-domain version by 0.86\%, demonstrating its effective use of cross-domain data to enhance downstream classification performance.

\noindent\textbf{Facial expression recognition ($\mathbb F$).}
The rights side of \cref{table1} shows the effectiveness of our DAP-MAE on facial expression recognition.
On both datasets, the experiments follow the same protocol, where 60 subjects are selected for 10-fold cross-validation. 
DAP-MAE achieves the highest recognition accuracy on BU-3DFE and Bosphorus compared with the latest facial expression recognition approaches. 
Our DAP-MAE outperforms the latest method DrFER \cite{drfer} by 0.66\% and 1.68\% on the BU-3DFE and Bosphorus datasets, respectively. Moreover, our method also surpasses most 2D+3D multi-modal approaches and achieves performance very close to the best-performing multi-modal method. As \cref{tab:my_label} shows, when pre-trained with simple combination of $\mathbb O+\mathbb F+\mathbb S$, Point-MAE and ReCon-SMC still perform lower than DAP-MAE and even lower than their same-domain version. This may due to the reason that other-domain data is treated as interference noises, as previously discussed.
\Cref{table1} shows that DAP-MAE not only performs well on coarse-grained classification tasks but also excels in fine-grained classification, highlighting its capability to address the domain gap problem.
%The above experiments showing that our method effectively captures both fine-grained and coarse-grained feature representations.
\begin{table}[!t]
    \centering
    \footnotesize
    \setlength{\tabcolsep}{3pt}
    \caption{Few-shot learning results on ModelNet40.%The method highlighted with \colorbox{cyan!8}{\phantom{X}} is pre-trained using $\mathbb O+\mathbb F+\mathbb S$. $\dagger$ denotes the reproduced results.
    } 
    \vspace{-5pt}
    \begin{tabular}{c c c c c }
        \hline
        \hline
        \multirow{2}{*}{Method}  &\multicolumn{2}{c}{5-way}&\multicolumn{2}{c}{10-way}\\
        \cline{2-3}
        \cline{4-5}
        &10-shot& 20-shot & 10-shot &20-shot \\
        \hline
        \multicolumn{5}{c}{\textit{Single-Modal Self-Supervised Learning}}\\
        \hline
        Point-BERT\cite{point-bert} $_\text{CVPR'22}$&94.6±3.1& 96.3±2.7& 91.0±5.4& 92.7±5.1  \\
        Point-MAE \cite{point-mae} $_\text{ECCV'22}$&96.3±2.5& 97.8±1.8& 92.6±4.1& 95.0±3.0  \\
        Point-M2AE \cite{point-m2ae} $_\text{NeruIPS'22}$& 96.8±1.8& 98.3±1.4 &92.3±4.5& 95.0±3.0\\
        PointGPT- \cite{pointgpt} $_\text{NeruIPS'23}$&96.8±2.0 &98.6±1.1 &92.6±4.6& 95.2±3.4\\
        Point-FEMAE\cite{point-femae} $_\text{AAAI'24}$&97.2±1.9 &98.6±1.3 &\textbf{94.0±3.3}& \textbf{95.8±2.8} \\
               % \cellcolor{cyan!8} ReCon-SMC$^\dagger$\cite{recon} (baseline) &\cellcolor{cyan!8}97.0±2.0&\cellcolor{cyan!8}98.6±1.2&\cellcolor{cyan!8}93.0±3.5&\cellcolor{cyan!8}95.0±2.8\\
        \textbf{DAP-MAE(Ours)} &\textbf{97.5±1.8}&\textbf{98.9±0.6}&93.3±3.9&95.2±2.8\\
        \hline
        \multicolumn{5}{c}{\textit{Cross-Modal Self-Supervised Learning}}\\
        \hline
        ACT\cite{act} $_\text{ICLR'23}$&96.8±2.3 &98.0±1.4& 93.3±4.0& 95.6±2.8\\
        Joint-MAE\cite{joint-mae}$_\text{IJCAI'23}$&96.7±2.2& 97.9±1.8& 92.6±3.7 &95.1±2.6\\
        I2P-MAE\cite{i2pmae} $_\text{CVPR'23}$&97.0±1.8& 98.3±1.3 &92.6±5.0& 95.5±3.0\\
        TAP\cite{tap} $_\text{ICCV'23}$&97.3±1.8& 97.8±1.9& 93.1±2.6 &95.8±1.0\\
        ReCon-full\cite{recon} $_\text{ICML'23}$&97.3±1.9 &98.9±1.2 &93.3±3.9 &95.8±3.0\\
        \hline
        \hline
    \end{tabular}

    \label{table2}
    \end{table}
    
\begin{table}[t]
    \centering
    \footnotesize 
    \setlength{\tabcolsep}{3pt}
\caption{Part segmentation results on the ShapeNetPart: Mean intersection over union for all classes $\text{mIoU}_c$ (\%) and all instances $\text{mIoU}_I$ (\%) for Part Segmentation. PM: pre-trained modality, \textit{PC}: point cloud, \textit{I}: image, \textit{T}: text. %The method highlighted with  \colorbox{cyan!8}{\phantom{X}} is pre-trained using $\mathbb O+\mathbb F+\mathbb S$. $\dagger$ denotes the reproduced results.
}  
    \vspace{-5pt}
    \begin{tabular}{ ccc c }
        \hline
        \hline
        Method &PM&$\text{mIoU}_c$& $\text{mIoU}_I$\\
        \hline
        \multicolumn{4}{c}{\textit{Supervised Learing}}\\
        \hline
        PointNet\cite{pointnet} $_\text{CVPR’17}$&\textit{PC}&80.4 &83.7\\
        PointNet++\cite{pointnet++} $_\text{NeruIPS’17}$&\textit{PC} &81.9 &85.1\\    
        PointMLP \cite{pointmlp} $_\text{ICLR’22}$&\textit{PC}& 84.6 &86.1\\
        \hline
        \multicolumn{4}{c}{\textit{Self-Supervised Learning}}\\
        \hline
        Transformer \cite{transformer} $_\text{NeruIPS’17}$&\textit{PC}&83.4 &84.7\\
        Transformer-OcCo \cite{occo} $_\text{ICCV’21}$&\textit{PC}&83.4 &85.1\\
        Point-BERT \cite{point-bert} $_\text{CVPR'22}$&\textit{PC}&84.1 &85.6  \\
        MaskPoint \cite{maskpoint} $_\text{ECCV’22}$&\textit{PC}&84.4 &86.0\\
        Point-MAE \cite{point-mae} $_\text{ECCV’22}$&\textit{PC}&84.2 &86.1  \\
        PointGPT-S\cite{pointgpt} $_\text{NeruIPS’23}$&\textit{PC}&84.1 &86.2 \\
        PM-MAE\cite{patchmixing} $_\text{TCSVT’24}$&\textit{PC}&84.3 &85.9\\
        Mamba3D+Point-MAE \cite{mamba3d} $_\text{ACM MM’24}$&\textit{PC} &83.6 &85.6\\
        %ReCon-SMC\cite{recon} $_\text{ICML’23}$ (baseline)&\textit{PC}&84.5 &86.1\\
        % \cellcolor{cyan!8} ReCon-SMC$^\dagger$\cite{recon} (baseline)&\cellcolor{cyan!8}\textit{PC}&\cellcolor{cyan!8}84.1 &\cellcolor{cyan!8}85.9\\
         \textbf{DAP-MAE(Ours)} &\textit{PC}&\textbf{84.9}&\textbf{86.3}\\
        \hline
        ACT\cite{act} $_\text{ICLR’23}$&\textit{PC+I}&84.7& 86.1\\
        ReCon-full\cite{recon} $_\text{ICML’23}$&\textit{PC+I+T}&84.9& 86.4\\
        \hline
        \hline
    \end{tabular}
    \label{table3}
    \vspace{-10pt}
    \end{table}

\noindent\textbf{Few-shot learning ($\mathbb O$).}
\Cref{table2} shows the results of few-shot learning, where each experiment undergoes 10 independent trials, and the average result is reported. Our DAP-MAE achieves the best performance among both single-modal and cross-modal methods, reaching 97.5\% on the ``5-way, 10-shot'' and 98.9\% on the ``5-way, 20-shot'' settings. For the ``10-way, 10-shot'' setting, our method outperforms all cross-modal methods and is only lower than Point-FEMAE. These results demonstrate that DAP-MAE, compared to other methods, maintains strong performance even with limited fine-tuning data.
\begin{table}[!t]
    \centering
    \footnotesize
    \setlength{\tabcolsep}{3pt}
    \caption{3D object detection on the ScanNetV2 dataset: The detection performance using Average Precision (AP) at two different IoU thresholds of 0.50 and 0.25, i.e., $\text{AP}_{50}$ and $\text{AP}_{25}$ are
reported. PM: pre-trained modality, \textit{PC}: point cloud, \textit{I}: image, \textit{T}: text. %The method highlighted with \colorbox{cyan!8}{\phantom{X}} is pre-trained using $\mathbb O+\mathbb F+\mathbb S$. $\dagger$ denotes the reproduced results.
}
    \vspace{-5pt}
    \begin{tabular}{c cc c c c}
        \hline
        \hline
        Method& PM&SSL& Input& $\text{AP}_{50}$ & $\text{AP}_{25}$\\
        \hline
        VoteNet\cite{votenet} $_\text{ICCV’19}$&\textit{PC}& $\times$ &xyz &33.5& 58.6\\
        PointContrast \cite{pointcontrast} $_\text{ECCV’20}$& \textit{PC}& \checkmark &xyz &38.0& 59.2\\
        STRL \cite{strl} $_\text{ICCV’21}$& \textit{PC}& \checkmark&xyz &38.4 &59.5\\
        RandomRooms \cite{randomrooms} $_\text{ICCV’21}$& \textit{PC}& \checkmark& xyz& 36.2& 61.3\\
        DepthContrast\cite{depthcontrast} $_\text{ICCV’21}$& \textit{PC}& \checkmark& xyz& - &61.3\\
        3DETR \cite{3detr} $_\text{ICCV’21}$&\textit{PC}& $\times$& xyz &37.9 &62.1\\
        Point-BERT \cite{point-bert} $_\text{CVPR’22}$& \textit{PC}& \checkmark& xyz &38.3& 61.0\\
        MaskPoint\cite{maskpoint} $_\text{ECCV’22}$& \textit{PC}& \checkmark&xyz &40.6& 63.4\\
        PiMAE\cite{pimae} $_\text{CVPR’23}$& \textit{PC+I}&\checkmark&xyz& 39.4&62.6\\
        ACT\cite{act} $_\text{ICLR’23}$&\textit{PC+I}& \checkmark &xyz& 42.1& 63.8\\
       % ReCon-SMC$^\dagger$\cite{recon} (baseline)&\textit{PC}&\checkmark&xyz&42.7&63.8\\
        % \cellcolor{cyan!8} ReCon-SMC$^\dagger$\cite{recon} (baseline)&\cellcolor{cyan!8}\textit{PC}&\cellcolor{cyan!8}\checkmark&\cellcolor{cyan!8}xyz&\cellcolor{cyan!8}42.5&\cellcolor{cyan!8}63.5\\
        \textbf{DAP-MAE(Ours)}&\textit{PC}&\checkmark&xyz&\textbf{43.2}&\textbf{64.0}\\
        \hline
        \hline
    \end{tabular} 

    \label{table4}
    \vspace{-10pt}
    \end{table}

\noindent\textbf{Part segmentation ($\mathbb O$).}
\Cref{table3} presents the results of object part segmentation on the ShapeNetPart \cite{shapenet}. 
Among all single-modal self-supervised methods, our DAP-MAE achieved the highest $\text{mIoU}_c$ and $\text{mIoU}_I$, reaching 84.9\% and 86.3\%, respectively, with $\text{mIoU}_c$ being 0.4\% higher than the runner up. 
Additionally, DAP-MAE outperformed existing supervised methods in the part segmentation task, being slightly lower than the cross-modal method ReCon-full only in the $\text{mIoU}_I$. This experiment demonstrates the effectiveness of our approach in segmentation tasks.

%Our DAP-MAE achieves the highest $\text{mIoU}_c$ of 84.9\% and $\text{mIoU}_I$ of 86.3\% among single-modal methods, surpassing the second-best $\text{mIoU}_c$ method ReCon-SMC by 0.4\% and outperforming the second-best $\text{mIoU}_I$ method PointGPT-S. Additionally, we outperform the best supervised method PointMLP, which surpasses all single-modal self-supervised methods in $\text{mIoU}_c$ except ours. Compared to cross-modal methods, DAP-MAE outperforms ACT \cite{act} and achieves performance comparable to ReCon-full.
    
\noindent\textbf{3D Object Detection ($\mathbb S$).}
DAP-MAE was evaluated on ScanNetV2 \cite{scannet} for the indoor 3D object detection, using the detection head of 3DETR \cite{3detr}. \Cref{table4} presents the Average Precision (AP) of different methods on the validation set, at thresholds 0.25 and 0.50.
Our DAP-MAE achieved 64.0\% $\text{AP}_{25}$ and 43.2\% $\text{AP}_{50}$, surpassing the baseline 3DETR by 5.3\% and 1.9\%, respectively, achieving the highest performance among all methods. 
Compared to PiMAE \cite{pimae}, a special method for 3D object detection, our DAP-MAE improves by 3.8\% and 1.4\% on the two metrics. 
Also, DAP-MAE outperforms the cross-modal approach ACT with gains of 1.1\% and 0.2\%, further demonstrating its effectiveness.
Furthermore, as shown in \cref{tab:my_label}, pre-training ReCon-SMC with cross-domain data leads to a 0.2\% $\text{AP}_{50}$ drop compared to the same-domain version, whereas DAP-MAE instead achieves a 0.5\% improvement.

\subsection{Ablation Study}
In this section, we conduct ablation studies for the cross-domain dataset (CD), heterogeneous domain adapter (HDA), and domain feature generator (DFG) to demonstrate their effectiveness in our DAP-MAE. 
All ablation studies are conducted on object classification on ScanObjectNN OBJ-BG. \Cref{table5} presents the ablation results.

As the left side of \cref{table5} shows, the baseline model pre-trained only on single-domain data, without CD, HDA, or DFG, achieved an accuracy of 94.15\% on the downstream object classification task. 
When combined with CD and pre-trained on a cross-domain dataset, the accuracy of the fine-tuned model was 94.32\%, which is only a slight improvement. If HDA or DFG is integrated, the model's accuracy can reach 94.66\%. This indicates that HDA collaboratively enhances the model's pre-training on cross-domain datasets, and DFG extracts domain features to guide the model's adaptation during fine-tuning for downstream tasks, both effectively improving the model's performance. When HDA and DFG are used simultaneously, the model's accuracy further increases to 95.18\%, which demonstrates that, as HDA and DFG play roles at different stages of DAP-MAE, they can collectively enhance the model's performance on the downstream task.

\begin{table}[!t]
    \centering
    \footnotesize
    \setlength{\tabcolsep}{3pt}% 压缩表格行间距% 调整表格横向位置，根据需要调整数值
            %\hspace*{-30pt}
    \caption{%
       Left: Ablation Study of DAP-MAE on ScanObjectNN (CD: cross-domain dataset, HDA: heterogeneous domain adapter, DFG: domain feature generator). Right top: The effect of fusion and w/o fusion in the HDA. Right bottom: Effect of using different fusion modes for varying domain combinations. 
    }
    \vspace{-5pt}
    \begin{tabular}{cc}
        % 左侧大表 (a)
    
        \begin{minipage}{0.4\linewidth}
        \renewcommand{\arraystretch}{1.2}
            \centering
            %\vspace{5pt}
            %\hspace*{-5pt}
            \footnotesize
            \begin{tabular}{cccc}
                \hline
                \hline
                CD & HDA & DFG & OBJ\_BG \\
                \hline
                $\times$ & $\times$ & $\times$ & 94.15 \\
                \checkmark & $\times$ & $\times$ & 94.32 \\
                \checkmark & \checkmark & $\times$ & 94.66 \\
                \checkmark & $\times$ & \checkmark & 94.66 \\
                \checkmark & \checkmark & \checkmark & \textbf{95.18} \\
                \hline
                \hline
            \end{tabular}
 % 调整标题与表格的间距
            %\vspace{3pt}
        \end{minipage}
        &
        % 右侧上下两个小表
        \begin{minipage}{0.5\linewidth}
            \centering
            \footnotesize
            %\hspace*{10pt}
            %\vspace{3pt}
     \begin{tabular}{ccc}
        \hline
        \hline
                 & \multicolumn{2}{c}{OBJ\_BG} \\ % 合并右侧两列并添加标题
        \hline
        w/o Fusion & \multicolumn{2}{c}{94.66} \\ % 合并右侧两列
        Fusion & \multicolumn{2}{c}{\textbf{95.18}} \\
        \hline
        \hline
        Fusion  Mode & ${\mathbb O}+{\mathbb F}$ & ${\mathbb O}+{\mathbb F}+ {\mathbb S}$ \\
        \hline
        Adding & 92.59 & 92.94 \\
        FC & 94.84 & 93.80 \\
        MLP & 93.80 & \textbf{95.18} \\
        \hline
        \hline
    \end{tabular}
        \end{minipage}
    \end{tabular}
    \vspace{-10pt}
    \label{table5}
\end{table}
The right side of \cref{table5} presents the ablation results of the fusion methods in HDA during the fine-tuning. 
As shown in the upper right of \cref{table5}, if HDA does not fuse the different domain data from its three MLP and only uses data from domain $\mathbb{O}$, the object classification accuracy is 94.66\%, which is 0.52\% lower than that (95.18\%) of the model with fusion. In contrast, the lower right of \cref{table5} shows the impact of different fusion modes in HDA on the model performance. One can observe that inappropriate fusion modes in HDA during the fine-tuning significantly reduce the classification accuracy. For instance, when data of the domain $\mathbb F$ and $\mathbb S$ are sequentially added to the data of domain $\mathbb O$ in HDA, the object classification accuracy drops from 94.66\% to 92.59\% and 92.94\%, respectively. If using a fully connected layer (FC) to predict the fusion coefficients, and sequentially fusing data of another two domains in HDA, object classification accuracies are 94.84\% and 93.80\%, respectively. When using MLPs to predict the fusion coefficients and fusing data from all three domains, the highest accuracy reaches 95.18\%. These results indicate that inappropriate fusion modes in HDA may cause the model to treat data from other domains as noise for the downstream tasks, thereby reducing their accuracy.

During the fine-tuning phase of DAP-MAE, the domain features $\bm d$ output by DFG, the class feature $\bm c$, and the point cloud feature $\mathcal F$ are input together into the downstream task head, guiding and enhancing its performance. The left side of \cref{table6} presents the results of different combinations of these three features on downstream task. It can be observed that the class feature $\bm c$, domain feature $\bm d$, and point cloud feature $\mathcal F$ can all be used independently for the downstream object classification. In the pre-training phase, the class feature $\bm c$ is trained in an unsupervised manner, resulting in a minimum accuracy of 93.12\% when used independently to fine-tune the object classification head. The domain feature $\bm d$ is supervised through contrastive learning during the pre-training, achieving a classification accuracy of 93.80\% when used independently to fine-tune the classification head. The point cloud feature $\mathcal F$ is supervised by reconstruction loss during the pre-training, which preserves complete point cloud information, and achieves an accuracy of 94.49\% when used to fine-tune the classification head. In the fine-tuning, combining these features generally improves the accuracy of the downstream classification, i.e. the highest accuracy reaches 95.18\% when all three features are used simultaneously.
%As shown in \cref{fig:3}, during the fine-tuning phase, we achieve the best results by feeding class feature $\bm c$, domain feature $\bm d$, and point cloud features $\mathcal F$ together into the downstream task head. In \cref{table:6}(b), we present the performance of different combinations of these three features. We can observe that using either $\bm c$ or $\bm d$, or even combining them, does not perform well. This indicates that the individual features decomposed from $\mathcal F$ are not enough for the downstream task.  However, when combined with $\mathcal F$, the result improves significantly. The combination of $\bm c$ and $\mathcal F$ achieves higher accuracy because class features are more suitable for the classification task. Adding $\bm d$ further adapts the model to the task, achieving the best accuracy.
\begin{table}[t]
    \centering
    \footnotesize
    \caption{Left: Classification performance of DAP-MAE with different feature combinations fed into the downstream task head. Right: Comparison of models' \#Params (M) and FLOPs (G).}\label{table6}

    \hspace*{5pt}
    \begin{minipage}[t]{0.3\linewidth}
        \centering
        \footnotesize
        %\caption*{(b)}
            \vspace{-1pt}
    \setlength{\tabcolsep}{.99mm}{
    \begin{tabular}{ccc|c}
        \hline
        \hline
        $\bm c$ & $\bm d/\bm o$ & $\mathcal F$ & OBJ\_BG \\
        \hline
        \checkmark & $\times$ & $\times$ &93.12 \\
        $\times$ & \checkmark & $\times$  &93.80\\
        $\times$ & $\times$ & \checkmark& 94.49\\
        \checkmark & \checkmark & $\times$ & 93.28\\
        \checkmark & $\times$ & \checkmark & 94.66\\
        $\times$ & \checkmark & \checkmark & 94.15\\
        \checkmark & \checkmark & \checkmark &\textbf{95.18} \\
        \hline
        \hline
    \end{tabular}}
     \end{minipage}
     \hspace{3pt}
    \begin{minipage}[t]{0.3\textwidth}
    \setlength{\tabcolsep}{2pt}
    \centering
    \footnotesize
    %\caption*{(c)}
    \vspace{-1pt}
    \setlength{\tabcolsep}{.99mm}{
    \begin{tabular}{c|cc}
        \hline
        \hline
        Method & \#P.(M) & FL. (G) \\
        \hline
        Point-MAE \cite{point-mae}  & 22.1  & 4.8  \\
        Point-M2AE \cite{point-m2ae}&15.3&3.6\\
        PointGPT-S \cite{pointgpt}     & 19.5  & -  \\
        ACT      \cite{act}       & 22.1  & 4.8 \\
        Point-FEMAE   \cite{point-femae}  & 27.4 & - \\
        ReCon \cite{recon} (baseline) & 43.6  & 5.3 \\
        DAP-MAE (ours)  & 43.8  & 5.4 \\
        \hline
        \hline
    \end{tabular}}
    \end{minipage}
    %\vspace{-10pt}
    \label{table:6}
    \vspace{-10pt}
\end{table}

The right side of \cref{table:6} compares the number of parameters (\#params) and computational complexity (FLOPs) across different methods. We use ReCon \cite{recon}, a dual-branch transformer with 43.6M parameters and 5.3G FLOPs, as the baseline because it enables us to better generate domain features. However, compared to ReCon, our DAP-MAE increases the number of parameters by only 0.2M and computational complexity by 0.1G, while achieving better performance on more downstream tasks. 
% \begin{wraptable}{r}{0.22\textwidth}
%     \centering
%     \footnotesize
%     \begin{tabular}{cc}
%     \hline
%     \hline
%       fusion mode   & OBJ\_BG \\
%      \hline
%        w/o fusion  &    94.66  \\
%          fusion   &     \textbf{95.18}\\
%     \hline
%     \hline
%     \end{tabular}
%     \caption{The effect of fusion and w/o fusion in the Dynamic Domain Adapter. }
%     \label{table7}
% \end{wraptable}
% \begin{table}[t]
%     \centering
%     \footnotesize
%     \renewcommand{\arraystretch}{1.2} % 调整行间距
%     \setlength{\tabcolsep}{4pt} % 调整列间距
%     \begin{tabular}{ccc}
%         \hline
%         \hline
%         Scale & $\mathbb O+\mathbb F$ & $\mathbb O+\mathbb F+ \mathbb S$ \\
%         \hline
%         1 & 92.59 & 92.94 \\
%         Linear & \textbf{94.84} & 93.80 \\
%         3-MLPs & 93.80 & \textbf{95.18} \\
%         \hline
%         \hline
%         \multicolumn{3}{c}{(c)}\\
%         \hline
%         \hline
%         Fusion Mode & \multicolumn{2}{c}{OBJ\_BG} \\ % 合并右侧两列并添加标题
%         \hline
%         w/o Fusion & \multicolumn{2}{c}{94.66} \\ % 合并右侧两列
%         Fusion & \multicolumn{2}{c}{\textbf{95.18}} \\
%         \hline
%         \hline
%         \multicolumn{3}{c}{(c)}\\
%     \end{tabular}
%     \caption{%
%         Combined results of Scale Generators and Fusion in DDA.
%     }
%     \label{table:combined}
% \end{table}
\section{Conclusion and limitation}
In this paper, we propose Domain-Adaptive Point Cloud Masked Autoencoder (DAP-MAE), a unified framework for general 3D point cloud analysis across multiple domains. With a single pre-training on a cross-domain point cloud dataset, DAP-MAE leverages a heterogeneous domain adapter and a domain feature generator to learn robust and transferable representations, enabling effective adaptation to diverse downstream tasks. Experimental results confirm the strong performance of DAP-MAE across various tasks and domains. However, the current model lacks the flexibility to incorporate new domains without retraining. In future work, we plan to explore continual learning strategies to enable domain expansion and improve cross-domain generalization, particularly in settings involving cross-modal frameworks.
%In the future, we plan to incorporate more domain data during pre-training to support a wider range of downstream tasks, aiming to surpass the performance of cross-modal methods.
%Theoretically, the performance of our DAP-MAE can continue to improve with the inclusion of additional domain datasets.
\section{Acknowledgements}
This work was supported by National Natural Science Foundation of China under Grant 82261138629, Guangdong-Macao Science and Technology Innovation Joint Fundation under Grant 2024A0505090003, Shenzhen Municipal Science and Technology Innovation Council under Grant JCYJ20220531101412030, and Guangdong Provincial Key Laboratory under Grant 2023B1212060076.
{
    \small
    \bibliographystyle{ieeenat_fullname}
    \bibliography{main}
}
\clearpage
\appendix
\setcounter{figure}{0}
\setcounter{table}{0}
\setcounter{equation}{0}
\setcounter{page}{1}
\twocolumn[{
\begin{center}
\section*{\Large DAP-MAE: Domain-Adaptive Point Cloud Masked Autoencoder for Effective Cross-Domain Learning}
{\large Supplementary Material}
%\vspace{-40pt}
\begin{figure}[H]
\hsize=\textwidth
        \centering
    \includegraphics[width=17.5cm]{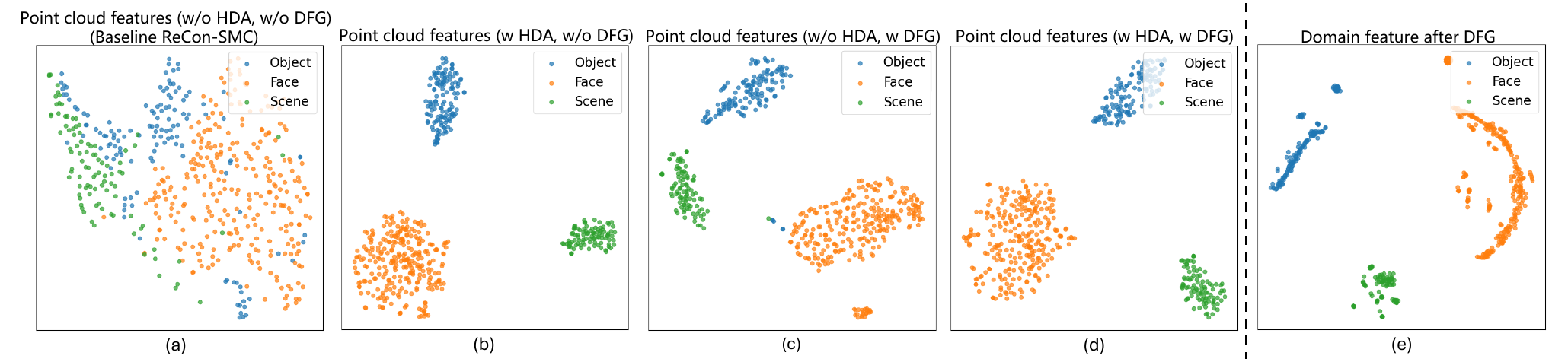}
    \caption{t-SNE visualization of features extracted from three domain datasets: (a) Point cloud features extracted w/o HDA and w/o DFG; (b) Point cloud features extracted w/ HDA and w/o DFG; (c) Point cloud features extracted w/o HDA and DFG; (d) Point cloud features extracted w/ HDA and w/ Domain DFG; (e) Domain features generated after DFG. }
    \label{fig:5}
\end{figure}
\end{center}

}]
\subsection{Visualization}
As shown in \cref{fig:5}, to better evaluate how DAP-MAE can collaboratively leverage cross-domain data to enhance the feature adaptability of the model and improve the performance of downstream tasks, we designed a t-SNE \cite{tsne} visualization experiment to assess the learned representations, which visualize the features extracted from the transformer encoder.

As shown in \cref{fig:5}(a), we first present the feature distribution of our baseline model, ReCon-SMC \cite{recon}, without applying the heterogeneous domain adapter (HDA) or integrating the domain feature generator (DFG), which is equivalent to the version pre-trained with simple combination of different domain data. The results show that the baseline has poor adaptability to the three different domains, as their features become intermingled, preventing the model from effectively learning and distinguishing each domain’s knowledge. This confusion can even mislead the model as noise, ultimately causing a drop in performance.
% we first visualize the features generated by our baseline model, ReCon-SMC, without applying the heterogeneous domain adapter (HDA) or integrating the domain feature generator (DFG). The results show that existing methods have limited feature adaptability across different domains, and cross-domain features often occupy the same feature space, indicating insufficient domain distinction.
\begin{figure}[h]
    \centering
    \includegraphics[width=0.8\linewidth]{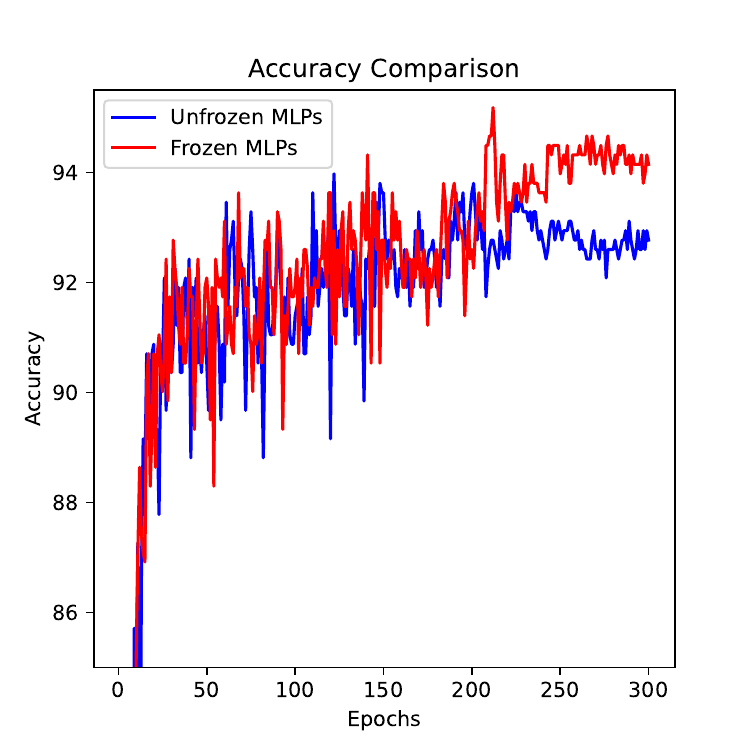}
    \caption{Comparison of classification accuracy during fine-tuning, showing how freezing or unfreezing the parameter of MLPs in HDA affects performance.}
    \label{fig:6}
\end{figure}
\cref{fig:5}(b) presents the features precessed with HDA. We can observe that features from different domains are well separated, with tight clustering within each domain. HDA processes data from different domains using separate MLP, creating distinct feature spaces and independently learning the point cloud geometry information for each domain. This indicates that the features no longer share the same feature space, allowing better adaptation for downstream tasks without interference from other domains. Furthermore, \cref{fig:5}(c) illustrates the features without HDA processing but concatenated with the domain feature generated by DFG. While the clustering performance is improved, some outliers still deviate from their domain centers. This situation may occur because the domain features learned by DFG come from the overall domain characteristics decomposed from point cloud features, which may not generalize well to individual samples. To further investigate, we independently visualize the domain feature in \cref{fig:5}(e), where similar issues of domain features deviating from their cluster centers are observed. However, we can see that each domain maintains its own distinctive distribution, indicating the model’s ability to learn unique domain feature patterns. When fine-tuned with tasks in the same domain, the model can leverage these patterns for rapid adaptation, leading to better performance. Finally, by combining the features processed through HDA and DFG, as shown in \cref{fig:5}(d), the clustering is significantly improved, demonstrating the effectiveness of our two contributions.
\subsection{Additional experiments}
\Cref{fig:6} compares classification accuracy during fine-tuning, highlighting the effects of freezing or not freezing the parameter of MLPs in HDA across fine-tuning epochs. Notably, in the final 100 epochs, the frozen MLPs approach achieves superior performance, while the unfrozen MLPs approach is prone to overfitting, resulting in a noticeable drop in accuracy.

The left side of \cref{table:7} illustrates the impact of loss function weights on the performance of downstream tasks. Our total loss function is defined as:
\begin{align}
    \mathcal{L}=w_1\mathcal{L}_{\text{rec}}+w_2\mathcal{L}_{\text{con}},
\end{align}
which consists of a reconstruction loss $\mathcal{L}_{\text{rec}}$ and a contrastive loss $\mathcal{L}_{\text{con}}$, balanced by weights $w_1$ and $w_2$ respectively. We can observe that increasing the weight of the reconstruction loss in the supervised setting while reducing the weight of the contrastive loss leads to better performance. This may be because the contrastive loss is prone to overfitting during pre-training.
\begin{table}[h]
    \centering
    \footnotesize
    \caption{Comparison of loss weight settings and learning rate effects.}
    \begin{minipage}{0.4\linewidth}
    \hspace*{15pt}
        \centering
        \begin{tabular}{cc|c}
            \hline
            \hline
            $w_1$ & $w_2$ & Accuracy \\
            \hline
            1.0  & 1.0  & 93.80 \\
            10  & 0.1  & 94.32 \\
            100   & 0.001  & \textbf{95.18} \\
            \hline
            \hline
        \end{tabular}
        \vspace{-1pt}
        \label{table:loss_weights}
    \end{minipage}
    \hfill
    \begin{minipage}{0.59\linewidth}
        \centering
        \begin{tabular}{c|c}
            \hline
            \hline
            Learning Rate & Accuracy  \\
            \hline
            0.001 & 94.84 \\
            0.0005 & \textbf{95.18} \\
            0.0001 & 94.84 \\
            \hline
            \hline
        \end{tabular}
        \label{table:lr_comparison}
    \end{minipage}
    \label{table:7}
\end{table}

\begin{figure}[t]
    \centering 
    \footnotesize
    \includegraphics[width=0.9\linewidth]{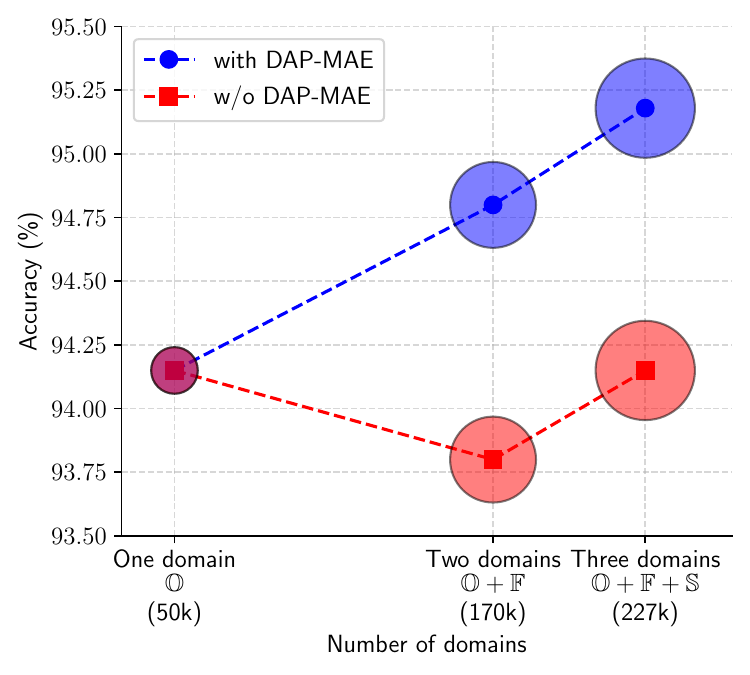}
    \caption{Effectiveness of cross-domain data.}
    \label{fig:2}
\end{figure}

\noindent\textbf{The effectiveness of cross-domain data.}
\Cref{fig:2} shows the object classification results of models pre-trained on point clouds from one, two, and three domains, with and without DAP-MAE. One can observe that simply increasing the data does not improve and may even decrease the object classification performance. In contrast, with DAP-MAE, as the training data gradually increases, the performance also gradually improves, showing no signs of saturation.

\begin{figure}[h]
    \centering 
    \footnotesize
    \includegraphics[width=1\linewidth]{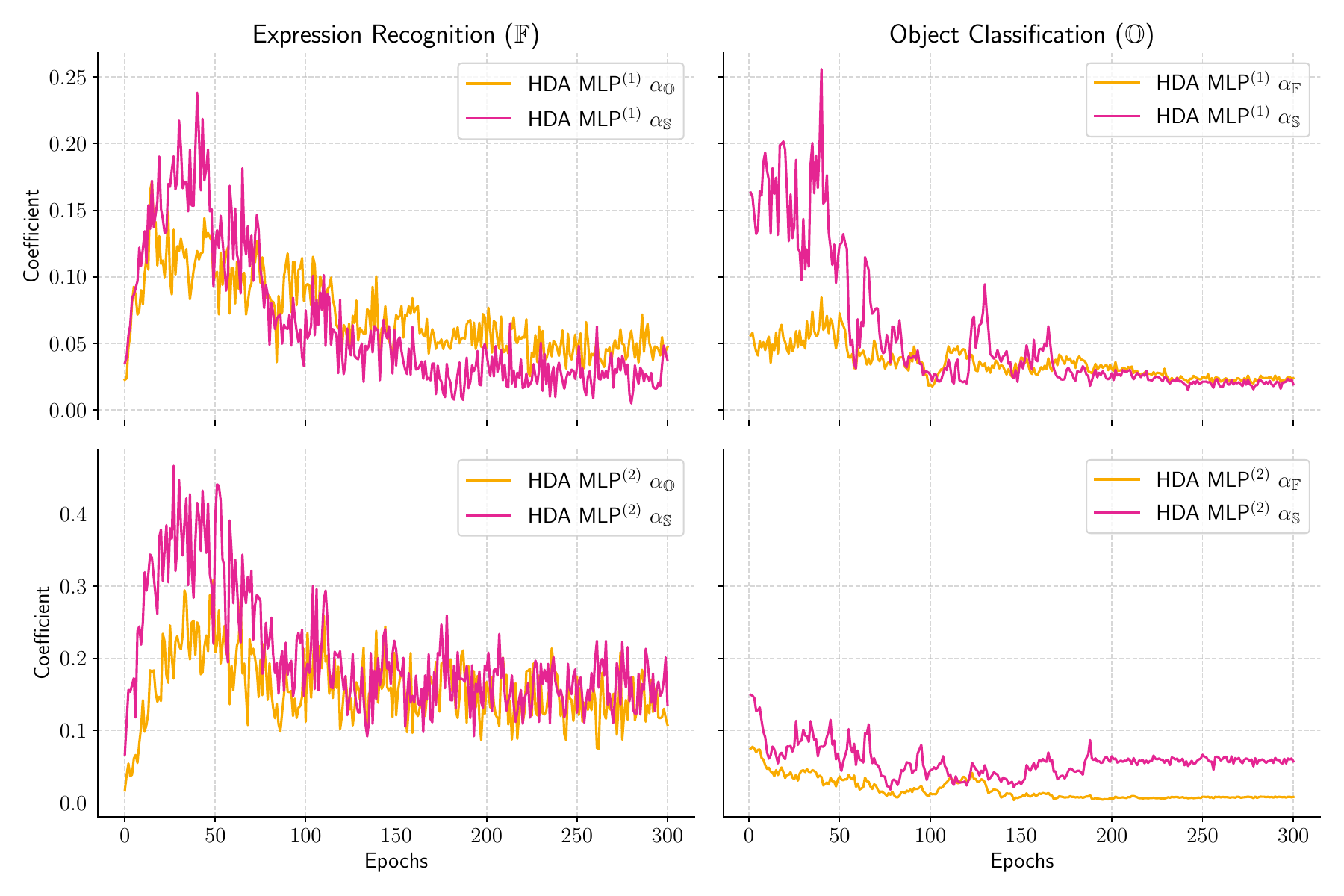}
    \caption{Coefficients learned by MLP in HDA.}
    \label{fig:1}
\end{figure}
\noindent\textbf{MLP coefficients.}
\Cref{fig:1} shows the evolution of the coefficients generated by the MLP$^{(1)}$ and MLP$^{(2)}$ during fine-tuning on the expression recognition and object classification. The coefficients all exhibit a trend of increasing followed by decreasing, indicating that in the early stages of fine-tuning, 
the learning capability of HDA on point clouds from other domains is more involved in the fine-tuning. As performance on the current task domain improves, they gradually withdraw.

\subsection{Experimental details.}
\noindent\textbf{Cross-domain dataset.}
ShapeNet \cite{shapenet} was captured from object ($\mathbb O$) domain, which contains more than 50,000 3D point clouds across 55 object categories. 
For the face domain ($\mathbb F$), the original FRGCv2 \cite{frgc} consists of 4,007 high-quality 3D face scans from 466 individuals with expression variations. In the pre-training, we utilized the enriched FRGCv2 \cite{gilani2018learning}, which contains about 120K 3D faces from 1K individuals. 
S3DIS \cite{s3dis} consists of six large-scale indoor scenes from three different buildings, covering a total of 273 million points across 13 categories. Only the training split of S3DIS was used. 
\begin{table*}[t]
    \centering
    \footnotesize
    \setlength{\tabcolsep}{5pt}
        \caption{Training details for different downstream tasks.}
    \begin{tabular}{lcccccc}
        \hline
        \hline
        Configuration & Object classification & Few-shot learning & Part segmentation &Facical expression recognition&Object detection\\
        \hline
        Optimizer  & AdamW & AdamW & AdamW&AdamW&AdamW \\
        Learning rate  & 5e-5 & 5e-4 & 8e-5 &1e-4&5e-5\\
        Batch size  & 32 & 32 & 64 &32&8\\
        Weight decay  & 0.05 & 0.05 & 0.05 &0.05&0.1\\
        Training epochs  & 300 & 150 & 300&300&1080 \\
        Warm-up epochs  & 10 & 10 & 10&0& 10\\
        Learning rate scheduler & Cosine & Cosine & Cosine &Cosine&Cosine\\
        Drop path rate & 0.1 & 0.1 & 0.2 &0.1&0.1\\
        \hline
        Number of points & 2048 & 1024 & 2048 &2048&40000\\
        Number of point patches  & 128 & 64 & 128&128&2048 \\
        Point patch sizes&32&32&32&32&64\\
        \hline
        \hline
    \end{tabular}
    \label{table:fine_tune_tasks}
\end{table*}

\noindent\textbf{Fine-tuning datasets.}
The pre-trained DAP-MAE was fine-tuned on five datasets, each corresponding to a different downstream task: object classification ($\mathbb{O}$), few-shot learning ($\mathbb{O}$), part segmentation ($\mathbb{O}$), facial expression recognition ($\mathbb{F}$), and 3D object detection ($\mathbb{S}$).

For object classification ($\mathbb O$), DAP-MAE was fine-tuned on ScanObjectNN \cite{scanojbectnn}, which consists of approximately 15,000 real-world objects across 15 diverse categories, and then evaluated using three different protocols, OBJ-BG, OBJ-ONLY, and PB-T50-RS. 
%Basic rotation data augmentation is used during fine-tuning. 
The few-shot learning ($\mathbb O$) experiments were conducted on the ModelNet40 \cite{modelnet} dataset, following the protocol established by \cite{point-mae,fewshot}. The experiments were structured as ``$n$-way, $m$-shot'', i.e. the training set contains $n$ selected categories and $m$ samples for each category $n \in \{5, 10\}$ and $m \in\{10, 20\}$.
%Each component undergoes 10 independent trials. 
Part segmentation ($\mathbb O$) was conducted on ShapeNetPart \cite{shapenet} which consists of 16,881 objects spanning 16 categories.

For facial expression recognition ($\mathbb F$), DAP-MAE was respectively fine-tuned on BU-3DFE \cite{bu3dfe} and Bosphorus \cite{bosphorus}. 
%BU-3DFE is a widely used publicly available dataset for academic research on 3D facial expression recognition. 
BU-3DFE contains 2,500 scans of 100 individuals (56 females and 44 males) aged between 18 and 70. Each individual has 25 samples representing seven different expressions: one neutral expression and six basic expressions with four different intensities. 
Bosphorus contains a total of 4,666 3D face scans collected from 105 individuals aged between 25 and 35. Among them, 65 individuals exhibit the six basic expressions with single intensity. 
%Both datasets follow the same experimental protocol, where 60 subjects are selected for 10-fold cross-validation, ensuring consistent evaluation across expressions.

For 3D object detection ($\mathbb S$), DAP-MAE was evaluated on ScanNetV2 \cite{scannet}, which consists of real-world richly annotated 3D point clouds of indoor scenes. ScanNetV2 comprises 1201 training scenes, 312 validation scenes, and 100 hidden test scenes. In ScanNetV2, 18 object categories are labeled using axis-aligned bounding boxes. 
%Our experimental results are evaluated on the validation set and we adopt 3DETR \cite{3detr} as the downstream task head.\

\noindent\textbf{Experiment setting.} As \cref{table:fine_tune_tasks} shows, the batch size was set to 512, and DAP-MAE was optimized using the AdamW optimizer with an initial learning rate of 0.0005 and a weight decay of 0.05. 
While the learning rate was decayed by a cosine schedule with a warm-up period of 10 epochs, the total epoch number was 300.
Random scaling and translation were used for data augmentation.
Also \cref{table:fine_tune_tasks} shows the fine-tune details on various downstream tasks. All experiments were conducted on an NVIDIA V100 GPU (32GB).

% {
%     \small
%     \bibliographystyle{ieeenat_fullname}
%     \bibliography{main}
% }

\end{document}